\documentclass[letterpaper]{article} 
\usepackage{aaai2026}  
\usepackage{times}  
\usepackage{helvet}  
\usepackage{courier}  
\usepackage[hyphens]{url}  
\usepackage{graphicx} 
\usepackage{natbib}  
\usepackage{caption} 
\usepackage{algorithm}
\usepackage{algorithmic}
\usepackage{amsmath}
\usepackage{amssymb}
\usepackage{colortbl}
\usepackage{booktabs}
\usepackage{newunicodechar}
\newunicodechar{≤}{\ensuremath{\leq}}

\usepackage{newfloat}
\usepackage{listings}
\DeclareCaptionStyle{ruled}{labelfont=normalfont,labelsep=colon,strut=off} 
\floatstyle{ruled}
\newfloat{listing}{tb}{lst}{}
\floatname{listing}{Listing}
\title{STAMP: Multi‑pattern Attention‑aware Multiple Instance Learning for STAS Diagnosis in Multi‑center Histopathology Images}
\author{
	Liangrui Pan\equalcontrib\textsuperscript{\rm 1},
	Xiaoyu Li\equalcontrib\textsuperscript{\rm 1},
	Guang Zhu\textsuperscript{\rm 1},
	Guanting Li\textsuperscript{\rm 1},
	Ruixin Wang\textsuperscript{\rm 1},
	Jiadi Luo\textsuperscript{\rm 2},
	Yaning Yang\textsuperscript{\rm 3}\protect\thanks{Corresponding author. Email: ynyang@xtu.edu.cn.},
	Liang Qingchun\textsuperscript{\rm 2}\protect\thanks{Corresponding author. Email: 503079@csu.edu.cn.},
	Shaoliang Peng\textsuperscript{\rm 1}\protect\thanks{Corresponding author. Email: slpeng@hnu.edu.cn.}
}
\affiliations{
	\textsuperscript{\rm 1}College of Computer Science and Electronic Engineering, Hunan University, Changsha 410082, China\\
	\textsuperscript{\rm 2}Department of Pathology, The Second Xiangya Hospital, Central South University, Changsha 410011, Hunan, China
	\textsuperscript{\rm 3}College of Computer Science and Electronic Engineering, Xiangtan University , Xiangtan 411000, China\\
}

\usepackage{bibentry}

\begin{document}

\maketitle

\begin{abstract}
	Spread through air spaces (STAS) constitutes a novel invasive pattern in lung adenocarcinoma (LUAD), associated with tumor recurrence and diminished survival rates. However, large-scale STAS diagnosis in LUAD remains a labor-intensive endeavor, compounded by the propensity for oversight and misdiagnosis due to its distinctive pathological characteristics and morphological features. Consequently, there is a pressing clinical imperative to leverage deep learning models for STAS diagnosis. This study initially assembled histopathological images from STAS patients at the Second Xiangya Hospital and the Third Xiangya  Hospital of Central South University, alongside the TCGA-LUAD cohort. Three senior pathologists conducted cross-verification annotations to construct the STAS-SXY, STAS-TXY, and STAS-TCGA datasets. We then propose a multi‑pattern attention-aware multiple instance learning framework, named STAMP, to analyze and diagnose the presence of STAS across multi‑center histopathology images. Specifically, the dual‑branch architecture guides the model to learn STAS‑associated pathological features from distinct semantic spaces. Transformer-based instance encoding and a multi‑pattern attention aggregation modules dynamically selects regions closely associated with STAS pathology, suppressing irrelevant noise and enhancing the discriminative power of global representations. Moreover, a similarity regularization constraint prevents feature redundancy across branches, thereby improving overall diagnostic accuracy. Extensive experiments demonstrated that STAMP achieved competitive diagnostic results on STAS-SXY, STAS-TXY and STAS-TCGA, with AUCs of 0.8058, 0.8017, and 0.7928, respectively, surpassing the clinical level. The 10 open baseline results establish a benchmark for STAS diagnostic research and facilitate the future generalizability and clinical integration of computational pathology technologies. Dataset features and code are accessible at \url{https://anonymous.4open.science/r/AAAI2026-3436}.
\end{abstract}


\section{Introduction}
\begin{figure}[ht]
	\centering
	\includegraphics[width=0.45\textwidth]{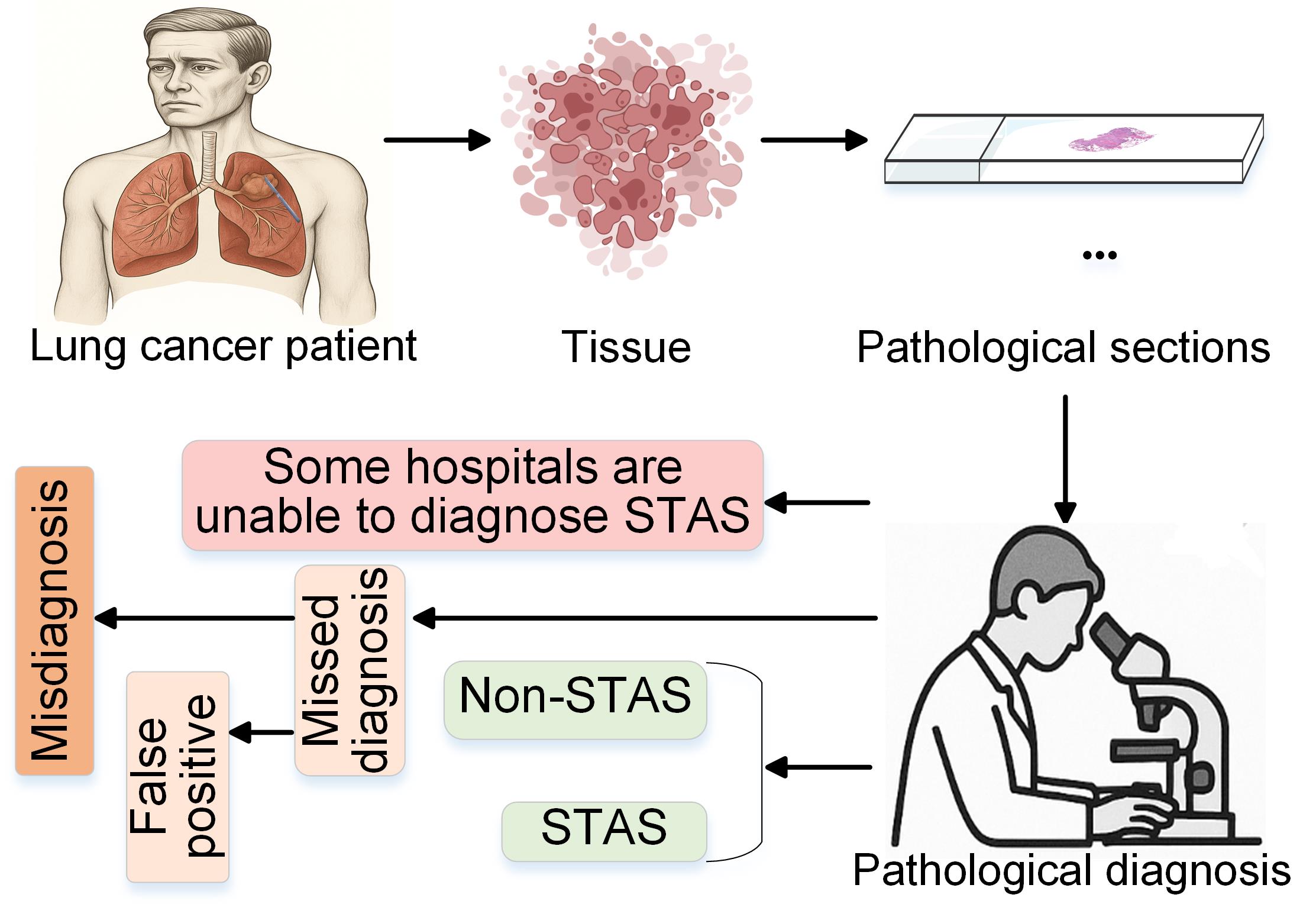}
	\caption{The challenges of STAS diagnosis include the unavailability of diagnostic results in some hospitals, a high risk of missed or incorrect diagnoses, and the labor-intensive nature of the diagnostic process.}
	\label{fig:data}
\end{figure}

Spread through air spaces (STAS) is a recently recognized pattern of invasion in lung cancer~\cite{han2021tumor, willner2024updates, ma2019spread, shih2020updates}. In 2013, Onozato et al.\cite{onozato2013tumor} first identified STAS using three-dimensional reconstruction, revealing that STAS consists of isolated tumor cells scattered in alveolar spaces, often separated from the primary tumor by several alveoli. In 2015, Kadota et al. formally described STAS in its current pathological context \cite{kadota2015tumor}. The World Health Organization later defined STAS as the presence of tumor cells such as micropapillary clusters, solid nests, or single cells within air spaces beyond the edge of the main tumor \cite{travis2015introduction}, and officially incorporated it into the pathological classification of lung adenocarcinoma (LUAD) \cite{travis2015introduction}. STAS is frequently observed in LUAD, with reported incidences exceeding 30\% in both domestic and international studies. It is significantly associated with aggressive clinicopathological features, such as pleural invasion, vascular invasion, larger tumor size, and advanced pathological stage \cite{han2021tumor, wang2019tumor, zhao2016clinicopathological,toyokawa2018significance}. Additionally, STAS is considered a strong predictor of occult lymph node metastasis in stage IA LUAD \cite{vaghjiani2020tumor}. Due to its association with poor prognosis and high recurrence risk after limited resection, STAS diagnosis has become a clinically essential task.

Histopathological images are considered the gold standard for diagnosing pulmonary STAS \cite{warth2015prognostic, zhou2022assessment}. As illustrated in Figure~\ref{fig:data}, the diagnosis of STAS heavily relies on pathologists performing detailed microscopic examinations of large quantities of histopathological slides. This process introduces substantial subjectivity into the diagnostic workflow. Moreover, hospitals with limited medical resources often lack the capacity to perform reliable STAS diagnosis \cite{villalba2021accuracy, li2024pro}. In addition, STAS diagnosis is a labor-intensive task that is prone to both misdiagnosis and missed diagnoses. A major cause of false-positive STAS results stems from nonstandard tissue sampling and slide preparation procedures, which may lead to the artificial detachment of tumor cells from the primary lesion \cite{villalba2021accuracy, ikeda2021current}. Intraoperative detection of STAS can guide surgical planning, including decisions between lobectomy and sublobar resection \cite{zhou2022assessment, villalba2021accuracy, ding2023value}. Postoperative STAS diagnosis also plays a crucial role in assisting oncologists in developing tailored treatment regimens. However, recent studies have reported that the AUC of manual STAS diagnosis ranges between 0.63 and 0.8, with an overall accuracy of approximately 0.7 \cite{villalba2021accuracy,zhou2022assessment}. Sensitivity tends to be relatively low, suggesting a high risk of missed diagnoses \cite{villalba2021accuracy,cao2025prediction}. Therefore, there is an urgent clinical need for an objective, accurate, and efficient method to assist in the diagnosis of STAS. 

\begin{figure*}[t]
	\centering
	\includegraphics[width=0.92\textwidth]{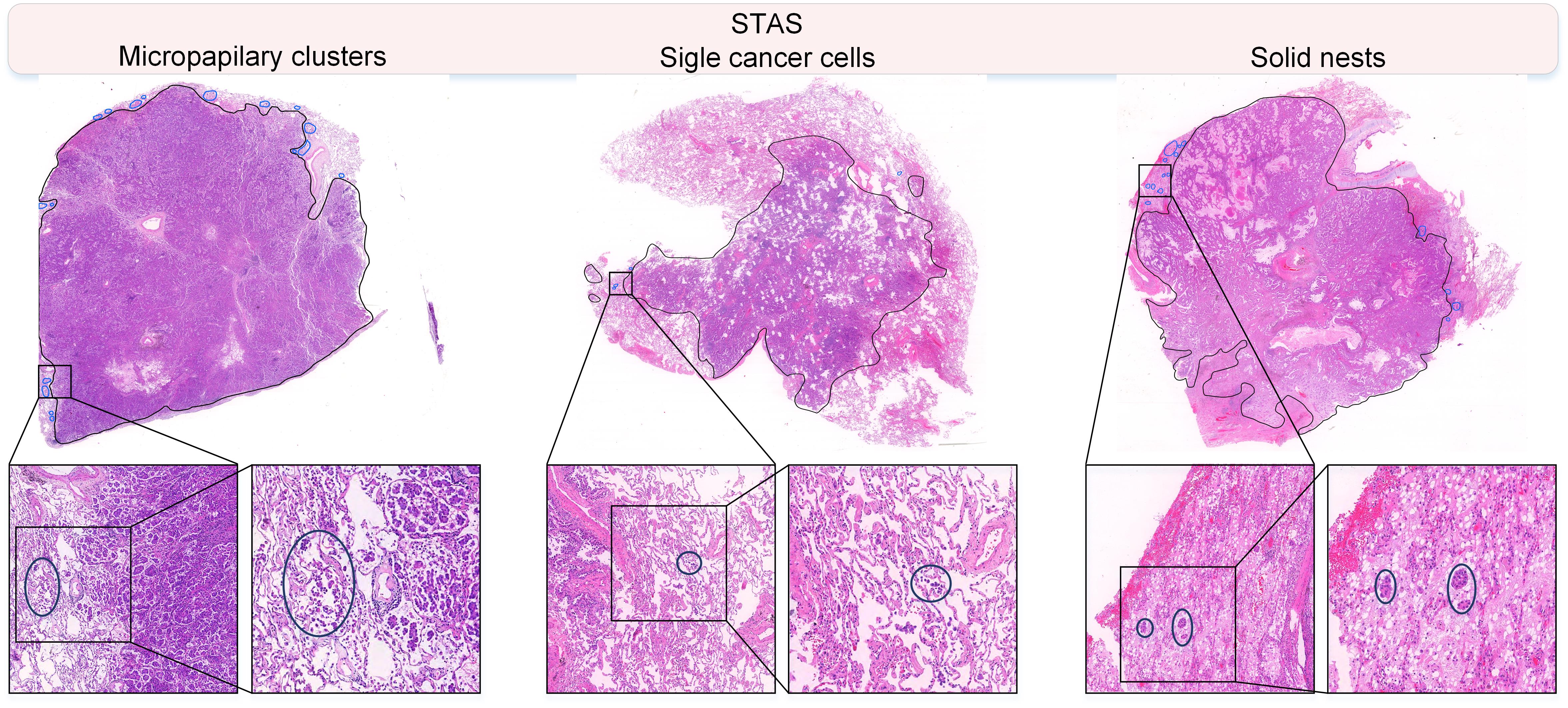}
	\caption{The three main pathological features of STAS in lung cancer histopathology images are as follows: STAS is mainly distributed outside the tumor body with the pathological features of micropapillary clusters, single cancer cells and solid nests.}
	\label{fig:stas}
\end{figure*}

With advances in deep learning, artificial intelligence (AI) has become increasingly integrated into pathological image analysis, yielding significant progress in cancer diagnosis and tissue grading~\cite{coudray2018classification,li2025application}. In light of the limitations of conventional pathology, AI-based approaches for detecting STAS in histopathological images are emerging as a promising direction. As illustrated in Figure~\ref{fig:stas}, STAS manifests in three histological forms: micropapillary clusters (lacking fibrovascular cores and occasionally forming ring-like structures within airways), solid nests (compact aggregates of tumor cells), and single tumor cells (individually scattered cells)~\cite{kadota2015tumor}. The morphological heterogeneity and subtlety of STAS features pose substantial challenges for AI models, particularly in accurately identifying irregular or sparse lesions. Moreover, STAS diagnosis requires high-quality whole-slide images (WSIs) and expert annotations, which are costly, labor-intensive, and inherently subjective, leading to limited and imbalanced training data. Variations in staining protocols and institutional differences further hinder model generalization. Despite the strong performance of deep learning models in standard image classification tasks, their accuracy in STAS detection on WSIs remains insufficient for clinical deployment. These challenges highlight the urgent need for more robust diagnostic strategies to improve the consistency, reliability, and efficiency of STAS identification.

To address the above challenges, we first collected histopathological images of STAS patients from the Second Xiangya Hospital and the Third Xiangya Hospital of Central South University, as well as the cancer genome atlas LUAD (TCGA\_LUAD) dataset. Three senior pathologists were invited to perform cross-validation annotations, resulting in the construction of the STAS-SXY, STAS-TXY, and STAS-TCGA datasets. Each slide was annotated for the presence or absence of STAS and the type of dissemination focus. Given the complex pathological features of STAS in WSIs, we propose a multi‑pattern attention-aware multiple instance learning (MIL) framework named STAMP for STAS diagnosis. The double-branch representation breaks the limitations of single-branch models by guiding the network to learn diverse lesion patterns from different semantic spaces, such as morphological features and contextual features, thereby enhancing feature diversity. Transformer-based instance encoding and a multi-pattern attention aggregation (MPAA) modules dynamically selects lesion-relevant key regions, avoiding noise interference and improving the discriminative power of the global representation. The similarity regularization constraint learning strategy prevents pattern redundancy, ensuring that the learned lesion patterns are highly discriminative and complementary. Our contributions are as follows:
\begin{itemize}
	\item We collaborated with three pathologists to construct and annotate three dedicated STAS datasets (STAS-SXY, STAS-TXY, and STAS-TCGA), which comprising a total of 2,011 WSIs. These datasets offer valuable resources for advancing clinical research on STAS in LUAD.
	
	\item To address the pathological complexity of STAS, we introduce STAMP, a multi‑pattern attention-aware MIL framework. Within the STAMP, the proposed MPAA module employs gated and content-aware dual-path attention at the feature level to dynamically highlight lesion-relevant regions, enabling the extraction of rich, heterogeneous features and enhancing WSI-level diagnostic accuracy.
	
	\item We conducted a extensive benchmark of 10 state-of-the-art MIL methods on the multi-center STAS datasets, providing comprehensive baseline results. Extensive ablation studies further validate the effectiveness and robustness of STAMP, highlighting its promise for advancing STAS-aware computational pathology.
	
\end{itemize}

\section{Related Work}
\subsubsection{Histopathology Image Analysis Methods}

Histopathological image classification has progressed through three major stages. Initially, traditional machine learning approaches relied on handcrafted features in conjunction with algorithms such as support vector machines, random forests, and k-nearest neighbors to classify WSIs \cite{cortes1995support, breiman2001random, zhang2016introduction}. With the emergence of deep learning around 2015, convolutional neural networks became dominant, enabling accurate classification of tumor subtypes and invasion levels across various cancer types, including breast, prostate, and colorectal cancers, which often achieving area under the receiver operating characteristic curves (AUCs) exceeding 0.8 \cite{pan2024deduce}. Since 2019, MIL has become the prevailing paradigm for WSI analysis due to its compatibility with weakly supervised learning scenarios. MIL methods can be broadly categorized into instance-based and bag-based approaches. Instance-based methods assign predictions to individual instances and then infer the bag-level label, assuming that a few key instances are sufficient to determine the final classification. Representative methods include MaxPooling, ABMIL \cite{ilse2018attention}, DSMIL~\cite{li2021dual}, ILRA~\cite{xiang2023exploring}, CLAM-SB, CLAM-MB~\cite{lu2021data}, and TransMIL~\cite{shao2021transmil}. In contrast, bag-based methods operate directly at the bag level, using similarity metrics (e.g., Gaussian kernel distances) without requiring instance-level labels. Examples include MeanPooling and WIKG~\cite{li2024dynamic}. Recent advancements integrating MIL with other methods, such as DeeMILIP, which embeds domain probability priors \cite{al2024incorporating}, MixMIL, which combines generalized linear mixed models with MIL \cite{engelmann2023mixed}, and cDP-MIL, which incorporates Bayesian nonparametrics for patch clustering \cite{chen2024cdp}, have further advanced MIL’s capabilities, enhancing both its efficiency and robustness in addressing challenges like adversarial example scarcity and uncertainty estimation.

\subsubsection{Diagnosis of STAS}
In recent years, only a few studies have focused on technical exploration and model innovation for the diagnosis of STAS \cite{pan2024feature}. Lee et al.\ analyzed 115 LUAD WSIs, generating over 600{,}000 H\&E patches to train a DL model via transfer learning, achieving an AUC of 0.738 for patch-level STAS detection \cite{lee2023ma20}. STASNet further improved semi-quantitative STAS identification across 489 WSIs, yielding a patch-level accuracy of 0.93 and a WSI-level AUC ranging from 0.72 to 0.78 \cite{feng2024deep}. It also computed STAS density and spatial distribution parameters, enhancing detection in real-time diagnostic workflows \cite{feng2024deep}. More recently, the VERN model employed a Siamese graph encoder to capture spatial-topological features, achieved high AUC in internal validation and AUCs of 0.8275 and 0.8829 in frozen and paraffin-embedded test sections, respectively, demonstrating clinical-grade performance \cite{pan2024feature}. Additionally, Gong et al.\ proposed a multi-field channel transformer (MFCT), which attained an F1-Score of 0.734 in STAS identification \cite{gong2025channel}. The SMILE framework introduced scale-aware MIL to adaptively select high-attention instances, diagnosing 251 and 319 STAS-positive WSIs on the CPTAC and TCGA datasets, respectively, which outperforming the average clinical AUC~\cite{pan2025smile}. Collectively, these studies demonstrate the promise of DL and MIL approaches in STAS diagnosis. However, existing methods still yield moderate AUCs, indicating a need for further advancement. To this end, we propose a a multi‑pattern attention-aware MIL framework, STAMP, to more effectively detect STAS from WSIs.

\section{Method}

\begin{figure*}[t]
	\centering
	\includegraphics[width=0.9\textwidth]{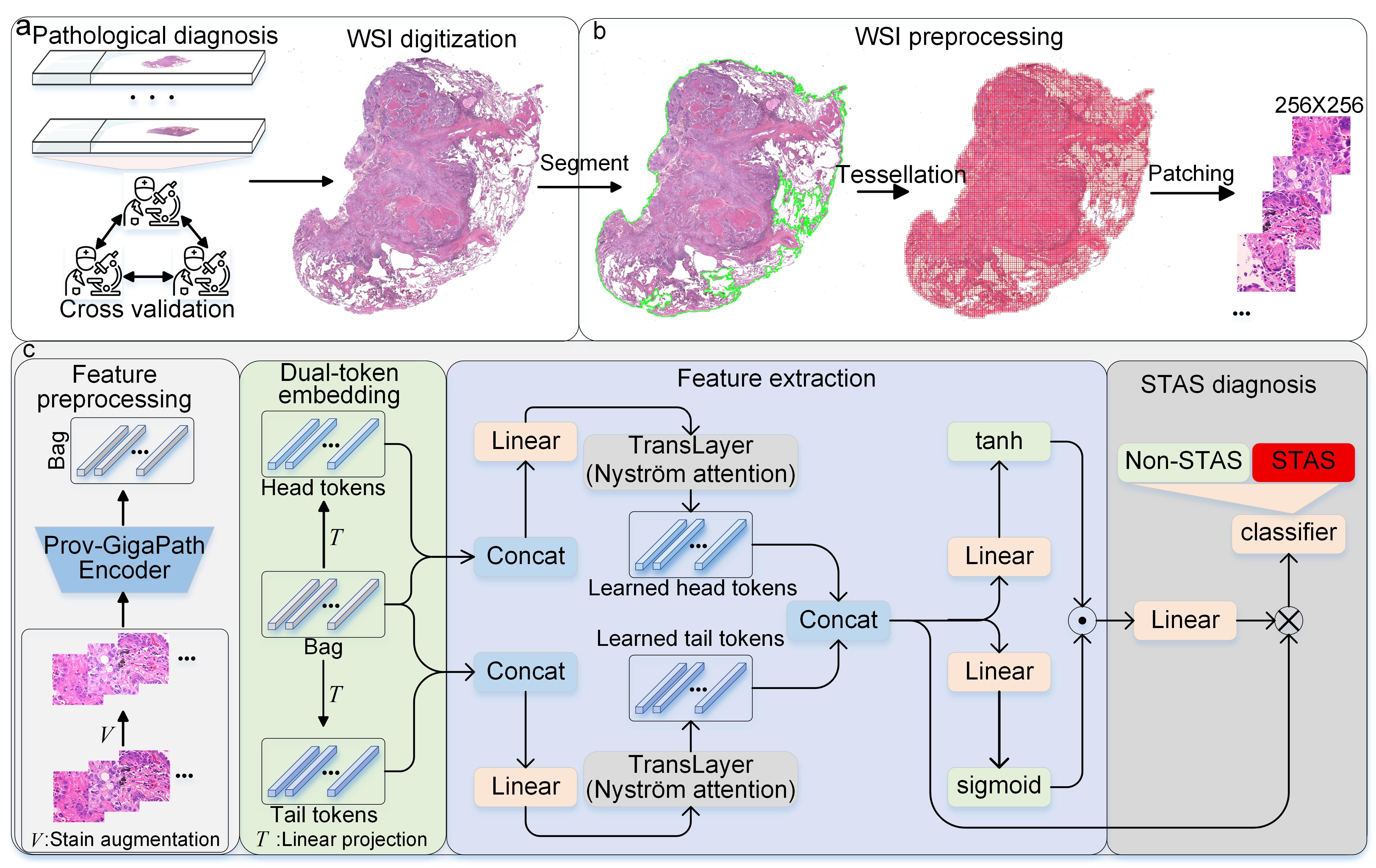}
	\caption{The complete workflow for diagnosing STAS from histopathological images. a): Annotation (cross-validation) of histopathological images from lung cancer patients and digitization of WSIs; b): Preprocessing of WSIs, including segmentation, tessellation, and patching; c): Preprocessing of WSI image features, dual-token embedding, feature extraction (including transformer-based instance encoding and MPAA modules), and STAS diagnosis (classification with regularized similarity loss).
	}
	\label{fig:folw}
\end{figure*}

\subsection{Notations}
In MIL, a WSI that has been partitioned into patches is represented as a bag $X = \{(x_1, y_1), \ldots,\ (x_n, y_n)\}$, where $x_i$ denotes the $i$-th patch (instance) in the bag, and $y_i$ represents its instance-level label (which is typically unknown in weakly supervised settings). The number of instances obtained from the WSI is denoted by $n$. Each bag $X$ is associated with a bag-level label $Y \in \{0, 1\}$, indicating whether the WSI contains a positive lesion region (e.g., STAS region). Under the weakly supervised MIL assumption, the relationship between the bag-level label and the instance-level labels is defined as follows:

\begin{equation}
	Y =
	\begin{cases}
		0, & \text{if } \sum y_i = 0\\
		1, & \text{otherwise } 
	\end{cases}
\end{equation}

In the STAS binary classification task, each instance \( x_i \) represents a patch from the histopathological image, and \( y_i \) denotes its corresponding label. The objective is to learn a classification model that can infer the label \( Y \) of a WSI based on its constituent patches, thereby identifying whether STAS-related pathological features are present in the WSI.

\subsection{Multi‑pattern Attention-aware Multiple Instance Learning}

\subsubsection{Overview}

In this section, we introduce STAMP, a novel multi‑pattern attention-aware MIL framework for weakly supervised WSI classification. Given a WSI represented as a bag of instances $X = \{x_1, x_2, \ldots, x_n\}\in \mathbb{R}^{n \times d}$, where each $x_i \in \mathbb{R}^d $ denotes a patch-level feature, STAMP aims to learn discriminative global lesion representations using a dual-token design and pattern-specific attention aggregation. The overall framework consists of four key components: (1) dual-token embedding, (2) Transformer-based instance encoding, (3) MPAA module, and (4) classification with regularized similarity loss.

\subsection{Dual-token Embedding}

To guide the learning of diverse lesion-aware representations, we introduce two sets of learnable tokens: a head token and a tail token, each consisting of \( n_p \) vectors initialized as parameters:
\begin{equation}
	\mathbf{T}_h, \mathbf{T}_t \in \mathbb{R}^{n_p \times d}
\end{equation}

These tokens are prepended to the original instance sequence and separately projected via two linear layers \( \mathbf{W}_h, \mathbf{W}_t \in \mathbb{R}^{d \times L} \) to obtain the head and tail branches:
\begin{equation}
	\mathbf{X}_h = \mathbf{W}_h([\mathbf{T}_h; X]), \quad \mathbf{X}_t = \mathbf{W}_t([\mathbf{T}_t; X])
\end{equation}
This double-branch design encourages the model to extract complementary semantic cues from distinct representation spaces.

\subsection{Transformer-based Instance Encoding}

Each token-augmented sequence \( \mathbf{X}_h \) and \( \mathbf{X}_t \) is independently fed into a lightweight Transformer encoder implemented via Nyström Attention \cite{xiong2021nystromformer} for efficiency and scalability. Specifically, each branch applies the following layer:
\begin{equation}
	\mathbf{H}_h = \text{TransLayer}(\mathbf{X}_h), \quad \mathbf{H}_t = \text{TransLayer}(\mathbf{X}_t)
\end{equation}
We retain the top \( n_p \) head and tail tokens from both branches and concatenate them as:
\begin{equation}
	\mathbf{H} = [\mathbf{H}_h^{(1:n_p)}; \mathbf{H}_t^{(1:n_p)}] \in \mathbb{R}^{2n_p \times L}
\end{equation}
These aggregated tokens are expected to encode region-specific semantic patterns useful for final classification.

\subsection{Multi-pattern Attention Aggregation}

To compute a global bag-level representation, we design a pattern-level attention module that combines gated and content-based attention:
\begin{align}
	\mathbf{A}_V &= \tanh(\mathbf{W}_v \mathbf{H}) \label{eq:av} \\
	\mathbf{A}_U &= \sigma(\mathbf{W}_u \mathbf{H}) \label{eq:au} \\
	\mathbf{A}   &= \text{softmax}\left(\mathbf{W}_a \left( \mathbf{A}_V \odot \mathbf{A}_U \right)\right) \label{eq:att}
\end{align}

Here, \( \odot \) denotes element-wise multiplication, and \( \mathbf{W}_v, \mathbf{W}_u \in \mathbb{R}^{L \times D} \), \( \mathbf{W}_a \in \mathbb{R}^{D \times 1} \) are trainable matrices. The attention weights \( \mathbf{A} \in \mathbb{R}^{1 \times 2n_p} \) are used to aggregate token features into a global representation \( \mathbf{M} \in \mathbb{R}^{1 \times L} \) via:
\begin{equation}
	\mathbf{M} = \mathbf{A} \cdot \mathbf{H}
\end{equation}
Where, $\cdot$ represents the product of vectors.

\subsection{Classification and Similarity Regularization}

The global feature \( \mathbf{M} \) is passed to a linear classifier for bag-level prediction:
\begin{equation}
	\hat{y} = \text{softmax}(\mathbf{W}_{\text{cls}} \mathbf{M})
\end{equation}
To enhance the diversity and orthogonality of the learned pattern tokens, we introduce a similarity regularization loss over the final token representations:
\begin{equation}
	\mathcal{L}_{\text{sim}} = \mathbb{E}_{i \neq j} \left[ \text{ReLU}(\cos(\mathbf{H}_i, \mathbf{H}_j)) \right]
\end{equation}
The total training loss combines standard cross-entropy with similarity regularization:
\begin{equation}
	\mathcal{L}_{\text{total}} = \lambda \cdot \mathcal{L}_{\text{CE}} + (1 - \lambda) \cdot \mathcal{L}_{\text{sim}}
\end{equation}
where \( \lambda \) is set to 0.9 by default.

\section{Experiments and Results}
\subsection{Datasets}

Given the propensity for misdiagnosis and underdiagnosis in STAS pathological assessment, three experienced pathologists employed a double‑blind protocol with cross‑validation to annotate each WSI for STAS status under microscopic examination, thereby ensuring accuracy and mitigating subjectivity, oversight and overdiagnosis. During the inclusion‑exclusion process, WSIs containing mechanical artifacts were precluded by the aforementioned pathologists with extensive diagnostic expertise. Although false positives in STAS differentiation are encountered in clinical pathology practice, we utilized pertinent immunohistochemical markers to label suspicious lesions, facilitating further discrimination between genuine STAS‑positive dissemination and histiocytic aggregates.

\noindent\textbf{STAS‑SXH}: We included WSIs from the Xiangya Second Hospital of Central South University cohort as our internal training and validation dataset. Among the 12,169 patients who underwent lung nodule resection and were histologically confirmed with LUAD between April 2020 and December 2023, 206 individuals with a definitive diagnosis of STAS and 150 individuals without STAS were selected. From these selected cases, we collected a total of 1,290 WSIs (approximately four slides per patient), along with associated clinical data and immunohistochemistry images (e.g., markers such as TTF‑1, CK, CD68) \cite{yatabe2002ttf,chistiakov2017cd68}.

\noindent\textbf{STAS‑TXH}: We included a cohort from the Third Xiangya Hospital of Central South University, consisting of 304 histopathology slides from 68 lung cancer patients diagnosed with STAS between 2022 and 2023, with each WSI annotated for the presence or absence of STAS.

\noindent\textbf{STAS‑TCGA}: We collected 417 WSIs from 366 patients in the TCGA\_LUAD cohort dataset based on the inclusion/exclusion method. All WSIs are accompanied by corresponding STAS labels.


\subsection{Experimental Details}
\begin{table*}[ht]
	\centering
	\resizebox{1\linewidth}{!}{
		\begin{tabular}{l rrrrr rrrrr rrrrr}
			\toprule
			Method & \multicolumn{5}{c}{STAS-SXY} & \multicolumn{5}{c}{STAS-TXY} & \multicolumn{5}{c}{STAS-TCGA} \\
			\cmidrule(lr){2-6} \cmidrule(lr){7-11} \cmidrule(lr){12-16}
			& ACC & AUC & Precision & Recall & F1-Score
			& ACC & AUC & Precision & Recall & F1-Score
			& ACC & AUC & Precision & Recall & F1-Score \\
			\midrule
			MaxPooling & 0.6894 & 0.7337 & 0.6988 & 0.6981 & 0.6842
			& 0.6693 & 0.6511 & 0.6803 & 0.6712 & 0.6274
			& 0.6137 & 0.6159 & 0.5995 & 0.6311 & 0.5606 \\
			MeanPooling& 0.7181 & 0.7733 & 0.7378 & 0.7253 & 0.7093
			& 0.7895 & 0.6926 & 0.6919 & 0.7139 & 0.7010
			& 0.5700 & 0.6069 & 0.6099 & 0.6343 & 0.5319 \\
			ABMIL      & 0.7061 & 0.7416 & 0.7142 & 0.7080 & 0.6965
			& 0.7428 & 0.7430 & 0.7262 & 0.7322 & 0.6996
			& 0.6236 & 0.6146 & 0.6283 & 0.6326 & 0.5602 \\
			DSMIL      & 0.7144 & 0.7666 & 0.7111 & 0.7163 & 0.7120
			& 0.7725 & 0.6954 & 0.7285 & 0.6813 & 0.6760
			& 0.5842 & 0.5630 & 0.5785 & 0.5955 & 0.5253 \\
			ILRA       & 0.7421 & 0.7828 & 0.7350 & 0.7359 & 0.7320
			& 0.7047 & 0.7621 & 0.7022 & 0.7355 & 0.6801
			& 0.5797 & 0.5964 & 0.6423 & 0.6316 & 0.5327 \\
			WKG       & 0.7227 & 0.7880 & 0.7376 & 0.7379 & 0.7196
			& 0.7567 & 0.7801 & 0.7386 & 0.7771 & 0.7289
			& 0.5654 & 0.6449 & 0.6411 & 0.6636 & 0.5312 \\
			Clam\_SB   & 0.7227 & 0.8042 & 0.7227 & 0.7227 & 0.6885
			& 0.7532 & \textbf{0.8130} & 0.7532 & 0.7532 & 0.6792
			& 0.7392 & 0.6097 & 0.7392 & 0.7392 & 0.5764 \\
			Clam\_MB   & 0.7328 & 0.7971 & 0.7328 & 0.7328 & 0.7214
			& 0.7606 & 0.7925 & 0.7606 & 0.7606 & 0.6910
			& \textbf{0.7464} & 0.6258 & \textbf{0.7464} & 0.7464 & 0.5337 \\
			TransMIL   & 0.7116 & 0.7382 & 0.7137 & 0.7072 & 0.7002
			& 0.7218 & 0.7796 & 0.7485 & 0.7385 & 0.6867
			& 0.5898 & 0.5872 & 0.6001 & 0.6181 & 0.5370 \\
			STAMP & \textbf{0.7449} & \textbf{0.8058} & \textbf{0.7394} & \textbf{0.7424} & \textbf{0.7376} & \textbf{0.8124} & 0.8017 & \textbf{0.7619} & \textbf{0.7785} & \textbf{0.7623} & 0.7186 & \textbf{0.7928} & 0.7197 & \textbf{0.7546} & \textbf{0.6962} \\
			\bottomrule
	\end{tabular}}
	\caption{Statistics of the results of diagnosing STAS in STAS-SXY, STAS-TXY, and STAS-TCGA based on 10 MIL methods.} 
	\label{tab:results_decimal}
\end{table*}

\subsubsection{Data Preprocessing.} 
The workflow for predicting STAS in WSIs using multi-center STAS datasets mainly consists of three steps (Figure~\ref{fig:folw}): (1) pathological diagnosis of STAS, (2) WSI preprocessing, and (3) feature preprocessing of WSIs. Specifically, pathologists first screen eligible pathological slides for STAS diagnosis according to inclusion and exclusion criteria, followed by digitization of the slides. WSI preprocessing mainly involves selecting regions of interest (ROIs) of lung cancer tissue and dividing the WSIs into patches. We use Otsu’s automatic thresholding method to segment the tissue, background, and blurred areas in the WSIs \cite{liu2009otsu}. Then, a sliding window strategy is applied to divide the 20$\times$ magnification WSIs into $256 \times 256$ patches while recording the spatial coordinates of each patch. To reduce the impact of variations caused by different scanning devices and slide quality on model generalization, we apply a structure-preserving GAN trained on TCGA to perform stain augmentation on all patches of the WSIs~\cite{wagner2023transformer, wagner2021structure}. Subsequently, all patches are processed using GigaPath for image feature extraction~\cite{janowczyk2019histoqc}.

\subsubsection{Training Details.}

We train our model using the Ranger optimizer, which combines RAdam and Lookahead for stable and efficient convergence. The initial learning rate is set to \( 1 \times 10^{-4} \), with a weight decay of \( 1 \times 10^{-5} \). A cosine annealing learning rate scheduler is adopted to gradually reduce the learning rate to a minimum value of \( 5 \times 10^{-6} \) over the training process.

We train the model for 50 epochs with a batch size of 1. The input feature dimension $d$ is set to 1536, and the hidden dimension $L$ is set to 512. To ensure robust performance evaluation, we conduct experiments with random seeds ranging from 0 to 5 and report the averaged results across these experiments.

\subsection{Results}

To assess the performance of MIL approaches in diagnosing STAS, we benchmarked ten representative models across three curated STAS datasets: MaxPooling, MeanPooling, ABMIL \cite{ilse2018attention}, DSMIL \cite{li2021dual}, ILRA \cite{xiang2023exploring}, WIKG \cite{lu2021data}, CLAM‑SB \cite{lu2021data}, CLAM‑MB \cite{lu2021data}, TransMIL \cite{shao2021transmil}, and STAMP. For each MIL method, we used accuracy (ACC), AUC, precision, recall, and F1-score to comprehensively evaluate the diagnostic performance of STAS. Below, we compare the performance of 10 MIL models on three STAS datasets.

In the STAS-SXY dataset, STAMP consistently outperformed all baselines, achieving the highest AUC (0.8058), F1-score (0.7376), and ACC (0.7449), demonstrating its superior capacity to detect STAS presence. Notably, ILRA, with an AUC of 0.7828, and CLAM-MB, attaining 0.7971, demonstrated strong performance, underscoring the effectiveness of multi-pattern attention aggregation and double branch framework. In contrast, simpler aggregation methods such as MaxPooling and MeanPooling showed limited effectiveness, underscoring their inability to capture critical histopathological cues.

On the STAS-TXY dataset, STAMP again achieved top performance, with the highest ACC (0.8124) and F1-score (0.7623). Although its AUC (0.8017) was slightly below that of CLAM-SB (0.8130), the results suggest comparable discriminative power. WIKG and CLAM-MB also performed competitively, indicating that bag-level modeling with clustering-aware attention is more robust for the diagnosis of STAS compared with instance-level methods such as DSMIL.

In the more challenging STAS-TCGA dataset, STAMP maintained strong performance, achieving the highest F1-score (0.6962) and AUC (0.7928) among all models. While CLAM-MB obtained the highest ACC (0.7464), its low F1-score (0.5337) reveals a trade-off between sensitivity and ACC, potentially caused by overfitting to dominant patterns. Traditional pooling strategies and early MIL models (e.g., ABMIL, DSMIL) suffered significant performance drops, with AUCs below 0.65 in most cases, reflecting their limitations in extracting complex pathological features of STAS.

\subsection{Ablation Study}
\subsubsection{Influence of Pattern Number.}

To assess the effect of the number of latent patterns on model performance, we evaluated variants of STAMP with different numbers of patterns (\( n_p \in \{1, 2, 3, 4, 5\} \)) on the STAS-SXY dataset. As summarized in Table~\ref{tab:pattern_number}, the model achieves the highest overall performance when \( n_p = 3 \), indicating an optimal trade-off between representational capacity and generalization. Fewer patterns (e.g., \( n_p = 2 \)) may lead to underfitting due to insufficient expressiveness, while a larger number (e.g., \( n_p = 4, 5 \)) introduces redundancy without performance gains. Based on these findings, we set \( n_p = 3 \) as the default configuration for all subsequent experiments.

\begin{table}[ht]
	\centering
	\resizebox{1\linewidth}{!}{\begin{tabular}{c ccccc}
			\toprule
			Pattern Number & ACC & AUC & Precision & Recall & F1-Score \\
			\midrule
			1 & 0.7264 & 0.7804 & 0.7297 & 0.7332 & 0.7219 \\
			2 & 0.7227 & 0.7827 & 0.7186 & 0.7232 & 0.7163 \\
			3 & \textbf{0.7449} & \textbf{0.7971} & \textbf{0.7394} & \textbf{0.7424} & \textbf{0.7376} \\
			4 & 0.7356 & 0.7917 & 0.7372 & 0.7418 & 0.7311 \\
			5 & 0.7329 & 0.7941 & 0.7391 & 0.7391 & 0.7271 \\
			\bottomrule
	\end{tabular}}
	\caption{Influence of Pattern Number on Model Performance.}
	\label{tab:pattern_number}
\end{table}

\subsubsection{Comparative Analysis of Single and Double-branches.}
\begin{figure}[ht]
	\centering
	\includegraphics[width=0.45\textwidth]{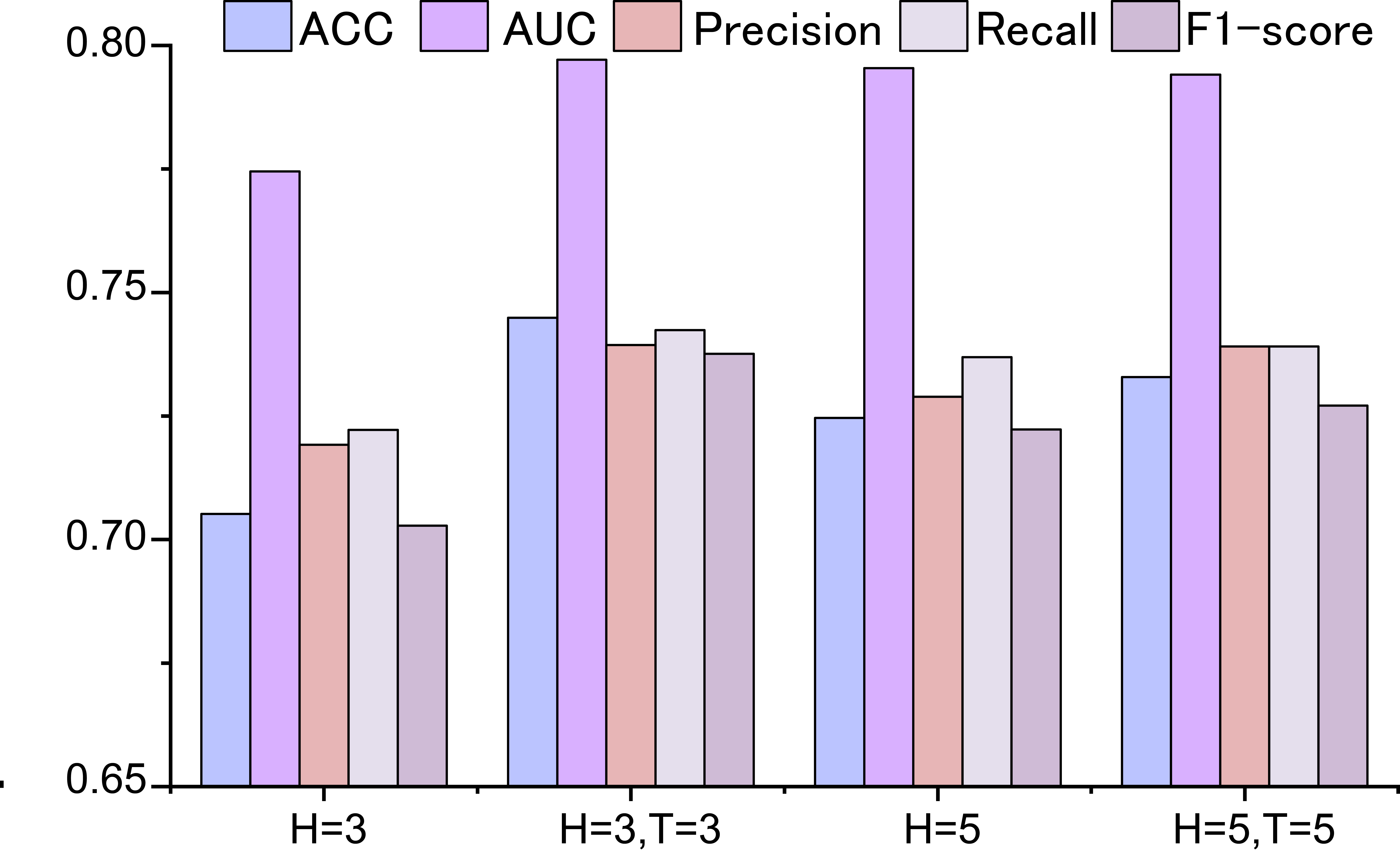}
	\caption{The impact of single-branch and double-branch structures on the performance of the STAMP.}
	\label{fig:fenzhi}
\end{figure}

To evaluate the effectiveness of the double-branch architecture, we conducted comparative experiments on STAS-SXY by varying the number of latent patterns in both single- and double-branch configurations. As shown in Figure~\ref{fig:fenzhi}, the double-branch (H,T) variant consistently outperforms its single-branch (H) counterpart across all metrics. For instance, with three latent patterns, the double-branch model achieves an accuracy of 0.7449, outperforming the single-branch version by 3.97\%. This improvement can be attributed to the enhanced feature representation enabled by the double-branch design. While the single-branch model is limited to a fixed semantic level and receptive field, the double-branch architecture leverages parallel sub-networks to extract complementary information across multiple semantic granularities and receptive fields. This results in a significantly enriched feature space, thereby enhancing both representational capacity and discriminative power.

\subsubsection{The Impact of Token Embedding Strategies.}

We further investigate the impact of instance feature embedding strategies on model performance. Specifically, we design two embedding variants: pre-embedding and post-embedding. The corresponding formulations are defined as follows:

\begin{equation}
	\small
	\begin{aligned}
		X_h = [W_h(X); T_h], \quad X_t = [W_t(X); T_t]
	\end{aligned}
\end{equation}

\begin{figure}[ht]
	\centering
	\includegraphics[width=0.45\textwidth]{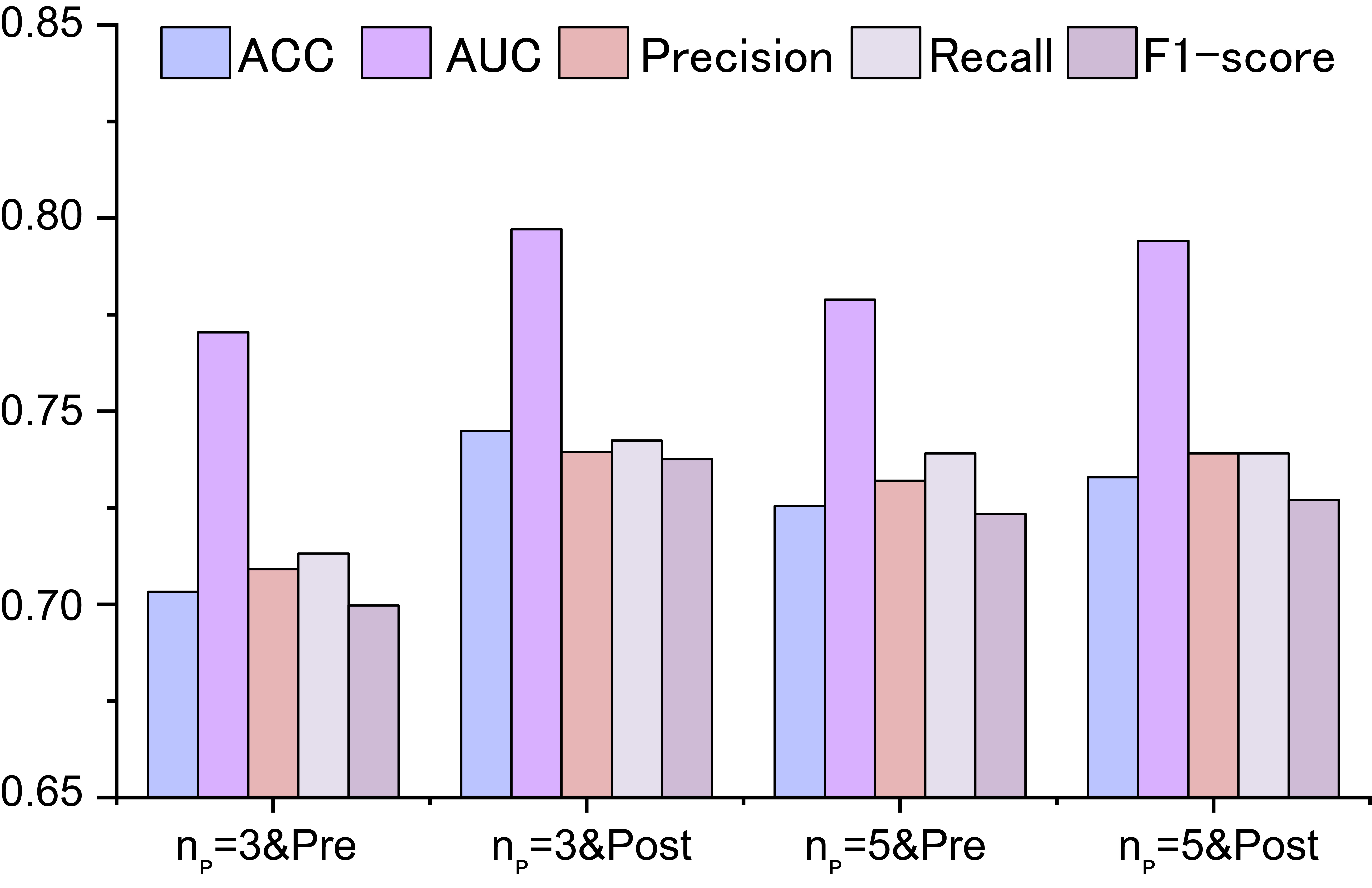}
	\caption{The impact of token embedding method on the STAMP performance. Pre and Post represent pre-embedding and post-embedding respectively.}
	\label{fig:token}
\end{figure}

As shown in Figure~\ref{fig:token}, across different numbers of latent patterns, the post-attention embedding consistently outperforms the pre-attention approach. This suggests that allowing the region-aware learnable tokens to interact with instance features at a later stage is more beneficial for the model, possibly due to enhanced contextualization and better alignment with the learned semantics.

\subsubsection{The Impact of Multi-pattern Attention Aggregation Strategies.}
To further evaluate the impact of multi-pattern attention aggregation strategies on model performance, we compare two different fusion methods: feature-level aggregation (FA) and prediction-level aggregation (PA). In our proposed STAMP framework, FA is adopted as the default setting. The PA strategy can be formulated as follows:

\begin{equation}
	\small
	\begin{aligned}
		M = \frac{1}{2} \Big[ & \text{softmax}(W_a^h \cdot (\tanh(W_v^h H_h^{(1:n_p)}) \odot \sigma(W_u^h H_h^{(1:n_p)}))) \\
		+ & \text{softmax}(W_a^t \cdot (\tanh(W_v^t H_t^{(1:n_p)}) \odot \sigma(W_u^t H_t^{(1:n_p)}))) \Big]
	\end{aligned}
\end{equation}

\begin{figure}[ht]
	\centering
	\includegraphics[width=0.45\textwidth]{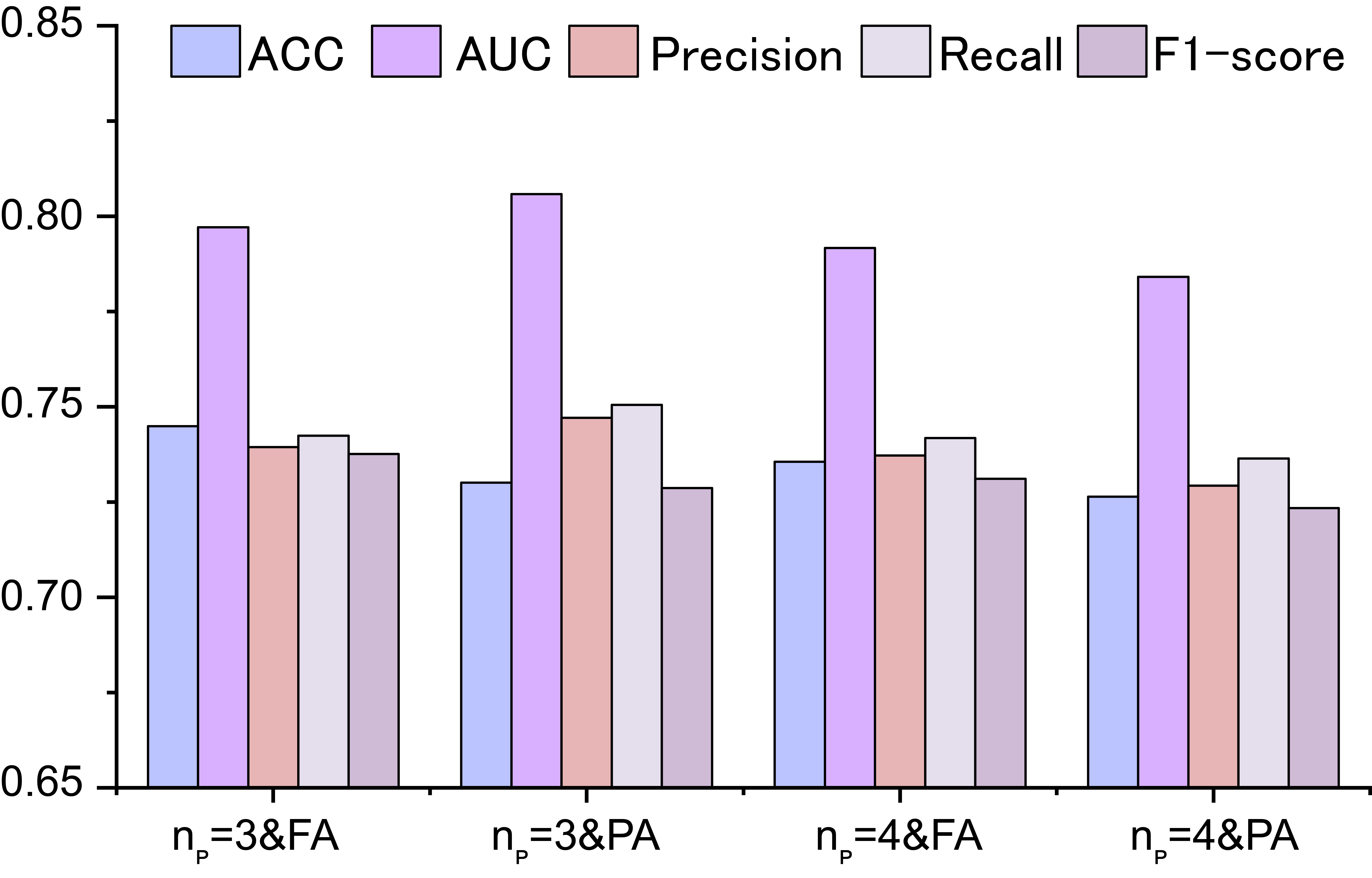}
	\caption{The impact of multi-pattern attention aggregation strategies on the STAMP performance.}
	\label{fig:fusion_strategy}
\end{figure}

As shown in Figure~\ref{fig:fusion_strategy}, FA consistently outperforms PA across different numbers of latent patterns. For instance, when $n_p=3$, FA achieves an accuracy of 0.7449, surpassing PA’s 0.7301. This result indicates that feature-level aggregation is more effective in enhancing the overall representational capacity and predictive performance of the model.

\section{Discussion and Conclusion}

In this study, we introduce three lung cancer STAS datasets to facilitate research on STAS diagnosis. To tackle the intricate pathological features of STAS, we propose STAMP, a multi‑pattern attention-aware MIL framework. We benchmark 10 MIL models on three datasets (STAS-SXY, STAS-TXY, and STAS-TCGA) for WSI-level STAS prediction. STAMP achieves state-of-the-art AUC and F1-score on STAS-SXY and STAS-TCGA, and superior accuracy and F1-score on STAS-TXY. Unlike existing MIL methods, the MPAA module in STAMP integrates gated and content-aware dual-path attention at the feature level, enabling adaptive focus on lesion-relevant regions and capturing diverse STAS-specific representations for improved WSI diagnostics. Ablation studies confirm that the double-branch architecture expands the feature space via parallel sub-networks for multi-scale and multi-granularity extraction, enhancing expressiveness and discriminability. Furthermore, post-instance feature embedding allows early interaction between learnable region-aware tokens and instance-level features, promoting nuanced STAS pattern learning, while feature-level multimodal fusion yields substantial performance gains. These findings underscore STAMP's potential for aiding clinical STAS detection in WSIs.


Despite these encouraging results, two primary limitations persist. First, the STAS datasets were curated in close collaboration with pathologists, with quality control reliant entirely on expert judgment, which a costly and labor-intensive process. Future work will involve collecting WSIs with common artifacts (e.g., mechanical distortions, bubbles, and staining inconsistencies) to train an automated quality control model for initial WSI screening. Second, while our model achieves strong performance on STAS-SXY and STAS-TCGA, it exhibits reduced efficacy on STAS-TXY, underscoring the need for enhanced cross-center generalization. Given the scarcity of STAS-positive cases, we plan to acquire additional positive WSIs to bolster training data. Moreover, although STAMP outperforms state-of-the-art MIL methods in diagnostic accuracy, bridging the disparity between model-derived attention maps and expert pathological annotations remains a key avenue for future research.

\bibliography{aaai2026}

\end{document}